\pdfoutput=1
\documentclass{article}

\usepackage[nonatbib,final]{iai_neurips_2024}

\usepackage[utf8]{inputenc} %
\usepackage[T1]{fontenc}    %
\usepackage{booktabs}       %
\usepackage{amsfonts}       %
\usepackage{nicefrac}       %
\usepackage{microtype}      %

\usepackage{amsmath, amssymb}
\usepackage{bbm}
\usepackage{dsfont}
\usepackage{xcolor}
\usepackage[bottom]{footmisc}
\usepackage{tabularx}
\usepackage{tablefootnote}
\usepackage[natbib]{biblatex}
\bibliography{_citations}
\usepackage[breaklinks]{hyperref}   %
\usepackage[noabbrev,capitalize]{cleveref}
\usepackage[bottom]{footmisc}
\usepackage{graphicx}
\usepackage{caption}
\usepackage{subcaption}
\usepackage{xspace}
\usepackage[normalem]{ulem}
\usepackage{multirow}
\usepackage{float}
\usepackage{ulem}
\usepackage{multicol}
\usepackage{enumitem}
\usepackage{cuted}
\usepackage{url}
\usepackage{setspace}
\usepackage{fancyhdr}
\usepackage{titlesec}

\crefformat{footnote}{#2\footnotemark[#1]#3}
\titlespacing*{\section}{0pt}{0.5ex plus 0ex minus 0ex}{0.25ex}
\titlespacing*{\subsection}{0pt}{0.5ex plus 0ex minus 0ex}{0.25ex}
\titlespacing*{\paragraph}{0pt}{0ex plus 0ex minus 0ex}{0.75em}

\usepackage{amsmath,amsfonts,bm}

\def\eqref#1{equation~\ref{#1}}

\def\1{\bm{1}}

\def\rb{{\textnormal{b}}}

\def\rvx{{\mathbf{x}}}
\def\rvy{{\mathbf{y}}}
\def\rvz{{\mathbf{z}}}

\def\rmH{{\mathbf{H}}}

\def\rmZ{{\mathbf{Z}}}

\DeclareMathAlphabet{\mathsfit}{\encodingdefault}{\sfdefault}{m}{sl}
\SetMathAlphabet{\mathsfit}{bold}{\encodingdefault}{\sfdefault}{bx}{n}

\def\gC{{\mathcal{C}}}

\def\gG{{\mathcal{G}}}

\def\gL{{\mathcal{L}}}

\def\gT{{\mathcal{T}}}

\def\gX{{\mathcal{X}}}
\def\gY{{\mathcal{Y}}}

\newcommand{\ifprecedingtext}[1]{\ifvmode\relax\else#1\fi}

\newcommand{\calm}{\textsc{CALM}\xspace}
\newcommand{\mask}{\texttt{[MASK]}\xspace}

\renewcommand{\rb}{RoBERTa\xspace}

\newcommand{\cls}{$g_Z$\xspace}
\newcommand{\hil}{$h_i^l$\xspace}

\def\ggt{{\mathcal{G}_{\mathcal{T}}}}
\def\ct{{\mathcal{C}_{\mathcal{T}}}}
\def\pa{{\mathbf{pa}}}

\DeclareMathOperator*{\topk}{top-\mathit{k}}
\DeclareMathOperator*{\doop}{do}
\DeclareMathOperator*{\val}{val}
\DeclareMathOperator*{\ovl}{overlap}
\DeclareMathOperator*{\acc}{acc}

\newenvironment{redenv}{
    \color{BrickRed}
}{
    \ignorespacesafterend
}

\newenvironment{blueenv}{
    \color{blue}
}{
    \ignorespacesafterend
}

\newenvironment{purpleenv}{
    \color{black}
}{
    \ignorespacesafterend
}

\newenvironment{oliveenv}{
    \color{olive}
}{
    \ignorespacesafterend
}

\newenvironment{tabenv}
   {\list{}{}%
    \item\relax}
   {\endlist}

\title{Competence-Based Analysis of Language Models}

\author{%
  Adam Davies \quad Jize Jiang \quad ChengXiang Zhai \\
  Siebel School of Computing and Data Science\\
  The Grainger College of Engineering\\
  University of Illinois Urbana-Champaign\\
  \texttt{adavies4@illinois.edu} \\
}

\begin{document}

\maketitle

\begin{abstract}
Despite the recent successes of large, pretrained neural language models (LLMs), comparatively little is known about the representations of linguistic structure they learn during pretraining, which can lead to unexpected behaviors in response to prompt variation or distribution shift.
To better understand these models and behaviors,
we introduce a general model analysis framework to 
study LLMs with respect to their representation and use of human-interpretable linguistic properties.
Our framework, \calm (Competence-based Analysis of Language Models), is designed to investigate LLM competence in the context of specific tasks
by intervening on models' internal representations of different linguistic properties 
using causal probing, and measuring models' alignment under these interventions with a given ground-truth causal model of the task.
We also develop a new approach for performing causal probing interventions using gradient-based adversarial attacks, which can target a broader range of properties and representations than prior techniques.
Finally, we carry out a case study of \calm using these interventions to analyze and compare LLM competence across a variety of lexical inference tasks,
showing that \calm can be used to explain behaviors across these tasks.
\end{abstract}

\section{Introduction}\label{sec:intro}
The rise of large, pretrained neural language models (LLMs) has led to rapid progress in a wide variety of natural language processing tasks 
\cite{gpt3,palm,dubey2024llama3}.
However, these models can also be quite sensitive to minor changes in input prompts \cite{pararel,notrobustinput,mizrahi2024multiprompt}
and fail to generalize outside their training or fine-tuning distribution \cite{wang2023chatgptrobustness,yang2023ood}.
It is usually unclear where these limitations come from, as LLM task performance is typically studied using only ``black box'' behavioral analysis
where limitations can only be detected if they are adequately represented in evaluation datasets, which cannot cover every potentially relevant limitation using a finite dataset \cite{everythingbenchmark,siska2024robustness}.
Thus, a deeper understanding of how these models can perform as well as they usually do while exhibiting unexpected limitations is critical for ensuring robust, trustworthy, and socially-responsible LLM-enabled applications 
\cite{shin2021explainableimplications,liao2023transparency,zou2023representation,bereska2024mechanistic}, 
and constitutes a key question in the basic science of LLM interpretation and analysis \cite{bereska2024mechanistic,anwar2024foundational}.

We approach this question in terms of \emph{competence}, drawing on the traditional competence-performance distinction in linguistic theory to motivate the study of LLMs in terms of their underlying representation of language.
We define LLM competence in the context a given linguistic task as the alignment between the ground-truth causal structure of the task and the LLM's latent representation of the task's structure, measured by intervening on the LLM's representation of task-causal versus spurious properties and observing how its behavior changes in response.
Models leveraging causal representations to perform a task generalize better under distribution shift than those that do not \citep{icp,irm,invariance}, meaning that more competent LLMs are also expected to exhibit greater robustness to distribution shift.

While the representations of causal or spurious properties are not directly observable,
we take inspiration from \emph{causal probing}, which intervenes on LLMs' representations of latent properties using causal interventions to study how these representations contribute to their behavior \cite{amnesic,causalgrammaticalnumber}.
We introduce a general model interpretation and analysis framework, \calm (for \emph{Competence-based Analysis of Language Models}), to study and measure LLM competence using causal probing.
While \calm can be instantiated using a variety of existing causal probing interventions (e.g., \citealp{inlp,kernel_inlp,ravfogel2022linear,spectral_nonlinear,belrose2024leace}), we develop a new methodology for intervening on LLM representations, \emph{Gradient-Based Interventions} (GBIs),
which use white-box adversarial attacks against supervised probes to modify LLM embedding representations. GBIs are the first causal probing technique that allow one to study models' use of arbitrarily-encoded feature representations,
enabling the investigation of new questions in language model interpretation. 
We carry out a case study of \calm 
using GBIs to intervene on two well-studied LLMs in order to measure and compare their competence across 14 lexical inference tasks, 
showing that \calm can indeed 
explain important patterns in behavior across these tasks by distinguishing between models' use of causal versus spurious properties.

\section{Competence-based Analysis of Language Models}\label{sec:calm}
\subsection{Linguistic Competence}\label{sec:competence}

Linguistic competence is generally understood as the ability to utilize one's knowledge of a language in producing and understanding utterances in that language, and is typically defined in contrast with linguistic performance, which is speakers' use of language in practice considered independently of the underlying knowledge that supports it \cite{marconi2020semanticcompetence}.\footnote{
    There has been significant debate in linguistics and the philosophy of language regarding the precise definition and nature of competence \cite{controversial2,controversial1,sag2011competence,marconi2020semanticcompetence}.
    However, the formalization of competence provided in this work is sufficiently general to incorporate most notions of competence, which may be flexibly specified by instantiating \calm in different ways.
}
Given a linguistic task, we may understand competence in terms of the underlying linguistic knowledge that one draws upon to perform the task.
If fluent human speakers rely on (implicit or explicit) knowledge of the same set of linguistic properties to perform a given task, then we may understand their performance of this task as being causally determined by these properties, and invariant to other properties.
For example, if we consider the two utterances ``the chicken crosses the road'' and ``the chickens cross the road'', the grammatical number of the subject (i.e., singular and plural, respectively) determines whether the verb ``(to) cross'' should be conjugated as ``crosses'' or ``cross''. 
As English (root) verb conjugation always depends on the grammatical number of the subject,
grammatical number may be regarded as having a causal role in the task of English verb conjugation, so we may understand fluent English speakers' mental representation of verb tense as having a causal role in their behavior.

In this work, we focus on \emph{lexicosemantic competence}, the ability to utilize knowledge of word meaning relationships in performing tasks such as lexical inference \cite{marconi1997lexicalcompetence,marconi2020semanticcompetence}.
While the study of human competence has a rich history in linguistics \cite{chomsky,controversial2,controversial1,sag2011competence,marconi1997lexicalcompetence,marconi2020semanticcompetence}, there is currently no generally accepted framework for studying LLM competence \cite{dissociating,pavlick2023symbols},
a gap which we aim to address in this work.

\subsection{\calm Framework}\label{sec:calmframework}

In order to make the study of competence tractable in the context of LLMs, 
we introduce the \calm (Competence-based Analysis of Language Models) framework, 
which describes an LLM's competence with respect to a given linguistic task in terms of its latent representation of the causal structure of the task.

\begin{figure}
\centering
\begin{minipage}[t]{.48\textwidth}
    \centering
    \includegraphics[width=\textwidth]{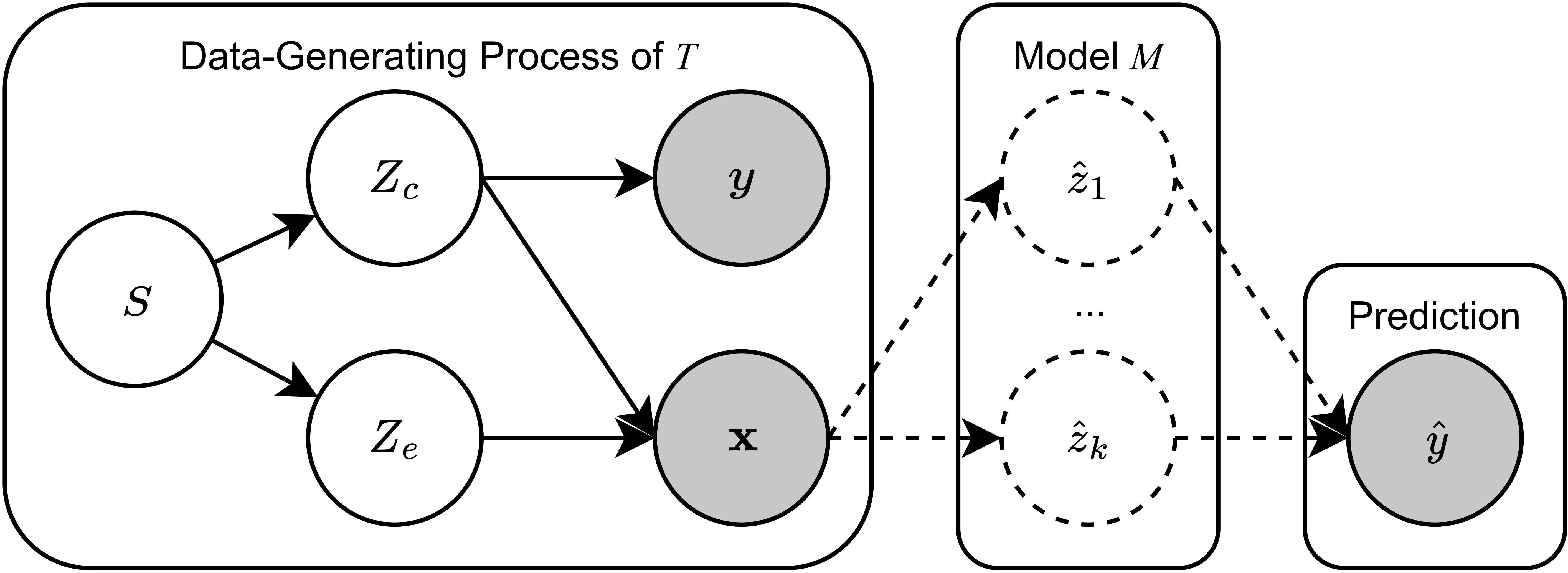}
    \caption{
    Structural causal model (SCM) of task $\gT$'s data-generating process and how it may be performed by model $M$. %
    Shaded and white nodes denote observed and unobserved variables, respectively. In \calm, the goal is to determine which representations $Z_j = z_j$ are causally implicated in $M$'s predictions $\hat{\mathbf{y}}$.
    }
    \label{fig:scm}
\end{minipage}%
\hspace{0.0125\textwidth}
\begin{minipage}[t]{.48\textwidth}
    \centering
    \includegraphics[width=\textwidth]{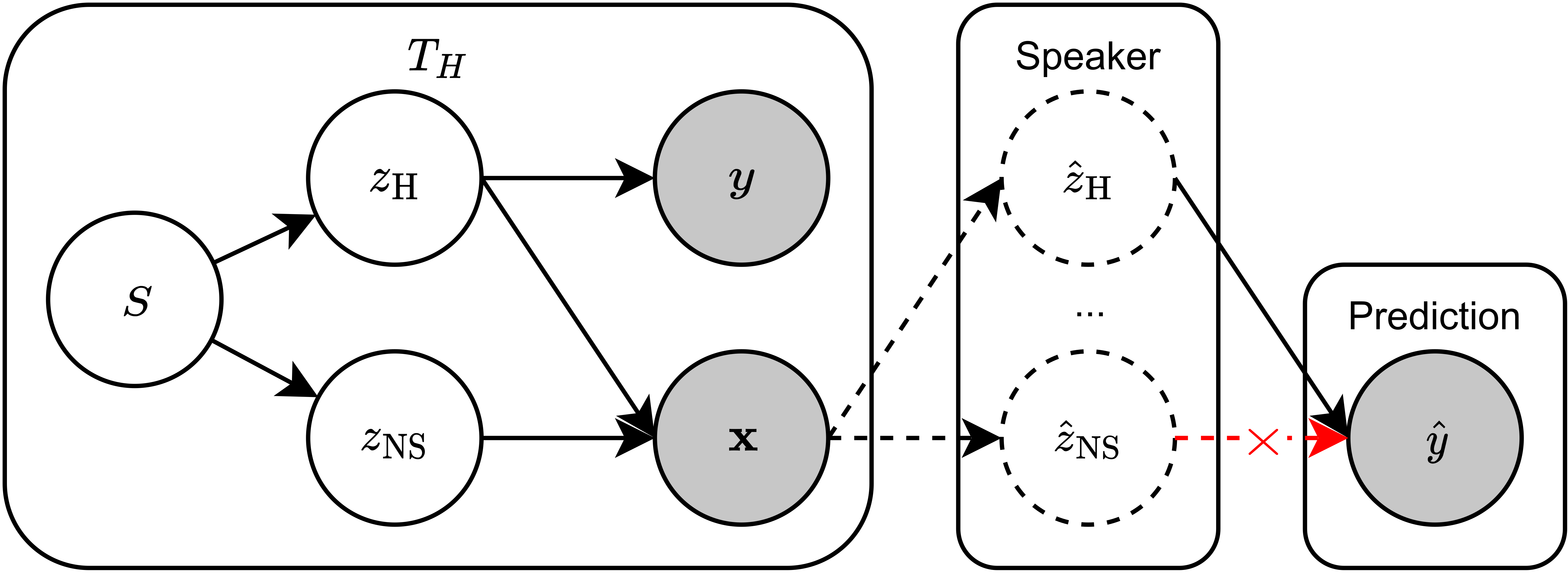}
    \caption{SCM of a competent English speaker on the hypernym prediction task. Shaded and white nodes denote observed and unobserved variables, respectively.}
    \label{fig:competent}
\end{minipage}
\end{figure}

\paragraph{Task Structure}
Formally, given supervised task $\gT \sim P(\mathcal{X}, \gY)$ where the goal is to correctly predict $\rvy \in \gY$ given $\mathbf{x} \in \mathcal{X}$, and a collection of latent properties $\rmZ = \{ Z_j \}_{j=1}^m$ that are (potentially) involved in generating $\mathbf{x}$, we formulate the causal structure of $\gT$ in terms of the data-generating process
\begin{equation}\label{eqn:dgp}
    \mathbf{x} \sim \Pr(\mathbf{x} | \rmZ_c, \rmZ_e), \quad \rvy \sim P(\rvy | \rmZ_c)
\end{equation}
where
$\rmZ$ may be decomposed into $\rmZ = \rmZ_c \cup \rmZ_e, \rmZ_c \cap \rmZ_e = \varnothing$, where $\rmZ_c$ contains all \emph{causal} properties that determine $\mathbf{y}$, and $\rmZ_e$ are the remaining (\emph{environmental}) properties that may be involved in generating $\mathbf{x}$ (cf. \citealp{scm}).
However, there may be an unobserved confounder $S$ that produces spurious correlations between $\rvy$ and $\rmZ_e$, which, if leveraged by language model $M$ in the course of predicting $\hat{\mathbf{y}}$, can lead to unexpected failures on $\gT$ when the spurious association is broken \cite{causaloverview}.
The structural causal model (SCM)\footnote{
    Note that an SCM is a directed acyclic graph where each node represents a variable and directed edges indicate causal dependencies -- see \citealt{bongers2021scm} for a brief overview.
} of this data-generating process is visualized on the left side of \cref{fig:scm}.

\label{rnd:orangutan}
For instance, suppose a speaker wants to communicate that orangutans are a genus of primate. She might say ``orangutans are primates'' or ``orangutans, a genus of apes, are primates''.
In both cases, the conjugation of the root verb would be ``are'' because the task of English verb conjugation (denoted $\gT_{\text{VC}}$) is invariant to whether the subject is complemented by an appositive phrase like ``a genus of apes'', and this phrase does not change the grammatical number of the subject ``orangutans''. 
Thus, if we define $Z_{\text{NS}}$ as the grammatical number of the subject, and $Z_{\text{AP}}$ as the presence of an appositive phrase modifying the subject, then $Z_{\text{NS}} \in \rmZ_c$ and $z_{\text{AP}} \in \rmZ_e$ for $\gT_{\text{VC}}$.
However, if we instead consider the task $\gT_{\text{H}}$ of predicting hypernyms -- for example, predicting $y$ in ``orangutans are $\mathbf{y}$s'', where $\mathbf{y} = $ ``primate'' and $\mathbf{y} = $ ``ape'' would both be correct answers -- 
then $Z_{\text{NS}} \in \rmZ_e$
(e.g., the same answers will be correct if the question is instead posed as ``an orangutan is a $\mathbf{y}$''), and the hypernymy relation $Z_{\text{H}}$ is instead be the causal property $Z_{\text{H}} \in \rmZ_c$, with the corresponding SCM visualized in \cref{fig:competent}.

\paragraph{Internal Representation}
Our main concern is measuring how attributable an LLM $M$'s behavior in a given task $\gT$ is to its representation of various properties $\rmZ = \{ Z_1, ..., Z_m \}$, and how these properties correspond to the causal structure of the task.
If $M$ respects the data-generating process of $\gT$, then its behavior should be attributable only to causal properties $Z \in \rmZ_c$ (and not to environmental properties $Z \in \rmZ_e$), in which case we say that $M$ is \emph{competent} with respect to $\gT$ (see \cref{fig:competent}).
We study model $M$'s use of each property $Z_j \in \rmZ$ by performing causal interventions $\doop(Z_j)$ on its representation of $Z_j$ in the course of performing task $\gT$, and measure the impact that these interventions have on its predictions.

\paragraph{Measuring Competence}\label{sec:metric}
We evaluate the competence of $M$ with respect to task $\gT \sim P(\gX, \gY)$ in terms of its relationship with a \emph{competence graph} $\ggt$,
which we define as a structural causal model (SCM) of $\gT$ with nodes corresponding to each latent variables $Z_j \in \rmZ$ and an additional node for outputs $\mathbf{y} \in \gY$
and directed edges denoting causal dependencies between these variables. 
That is, the set of causal properties $\rmZ_c$ defined by $\ggt$ is the set of all properties $Z_j \in \rmZ$ such that there is an edge or path from $Z_j$ to $\mathbf{y}$.

To study the extent to which $M$'s behavior is determined by causal versus spurious dependencies in $\ggt$, we examine whether $M$ and $\ggt$ make the same predictions under interventions $\doop(\rvz)$, where 
setting $\rvz = \{ z_j \}_{j=1}^{m}$ 
is the set of values 
$Z_j = z_j$ 
taken by each corresponding latent variable $Z_j \in \rmZ$.
For example, consider an instance $(\rvx, \rvy) \sim \gT_H$ of the hypernym prediction task $\gT_H$ where input $\rvx =$``orangutans are $\mathbf{y}$s'' and ground-truth output $\mathbf{y} =$``primate''.
Here, the values taken by $\rvz$ would be $Z_{\text{H}} = 1, Z_{\text{NS}} = 1$ (where $1$ indicates the presence of hypernymy and a plural noun subject, respectively),
and we might define an alternative $\rvz'$ where $Z_{\text{H}} = 0, Z_{\text{NS}} = 1$, under which a competent model's prediction would be expected to change with the causal variable $Z_{\text{H}}$ (i.e., $M(\mathbf{x} | \doop(\rvz')) \neq M(\mathbf{x})$).

The alignment of $M$ with $\ggt$ is measured in terms of the similarity $S$ of their predictions under interventions $\doop(\rvz)$ given input $\rvx \sim P(\gX)$,
and can be computed using a given similarity metric $S: \gY, \gY \to [0, 1]$ (e.g., equality, %
n-gram overlap, cosine similarity,
etc.) depending on the SCM $\ggt$ and output space $\gY$.
That is, we define $\ct(M | \ggt)$ as $M$'s competence with respect to task $\gT$ as a function of its alignment with corresponding task SCM $\ggt$ under interventions $\doop(\rvz)$ measured by similarity metric $S$, as follows:
\begin{equation}\label{eqn:competence}
    \ct(M | \ggt) = \mathbb{E}_{\rvx, \rvz \sim P(\gX, \val(\rmZ))} S \big( M (\rvx | \doop(\rvz)), \ggt (\rvx | \doop(\rvz)) \big)
\end{equation}
This $\ct(M | \ggt)$ metric (bounded by $[0, 1]$) is an adaptation of the Interchange Intervention Accuracy (IIA) metric \citep{iia,geiger2023causalinterp} to the context of causal probing, where instance-level interventions are replaced with concept-level interventions enabled by the gradient-based intervention methodology we introduce in \cref{sec:gbi}. (See \cref{apx:iia} for a detailed comparison of our competence metric with IIA.)

\paragraph{Causal Probing}\label{sec:causal}
A key technical challenge in implementing \calm (and causal probing more generally) is designing an algorithm to perform causal interventions $\doop(Z)$ that maximally damage the representation of a property $Z$ while otherwise minimally damaging representations of other properties $Z^\prime$ \cite{canby2024cpr,kernel_inlp} (see \cref{apx:gbi_limitations}).
For example, \emph{amnesic probing} \cite{amnesic} uses the INLP algorithm \cite{inlp} to produce interventions $g_{Z}$ that remove all information that is linearly predictive of property $Z$ from a pre-computed set of embedding representations $\rmH$,
showing that BERT makes variable use of parts-of-speech, syntactic dependencies, and named-entity types in performing masked language modeling.
However, \citet{amnesic} also found that, when INLP is used to remove BERT's representation of these properties in early layers, it is often able to ``recover'' this representation in later layers, which is likely due to BERT encoding these properties nonlinearly; and later work has 
found that the same ``recoverability'' problem persists even when linear information removal methods like INLP are kernelized \cite{kernel_inlp}.
Thus, it is necessary to develop interventions that do not require restrictive assumptions about the structure of LLMs' representations (e.g., linearity), %
a problem which we aim to solve in the following section.

\section{Gradient-Based Interventions}\label{sec:gbi}

Our goal in developing gradient-based interventions (GBIs) as a causal probing technique is to enable interventions over arbitrarily-encoded LLM representations.
GBIs allow users to flexibly specify the class of representations they wish to target, expanding the scope of causal probing to arbitrarily-encoded properties.
We take inspiration from \citet{latentattack}, who developed a technique to perturb latent representations using gradient-based adversarial attacks.\footnote{
    Notably, \citet{earlygbi} developed a similar methodology without explicit use of such attacks (see \cref{sec:relwk}).
}
They begin by training probe \cls$: \mathbf{h} \mapsto z$ to predict image class $z \in Z$ from latent representations $\mathbf{h} = f_{\text{enc}}(\mathbf{x})$ of images $\mathbf{x}$, where $f_{\text{enc}}$ is the encoder of a VAE-GAN \cite{vaegan} trained on an unsupervised image reconstruction task (i.e., $f_{\text{dec}}(f_{\text{enc}}(\mathbf{x})) = \hat{\mathbf{x}} \approx \mathbf{x}$, for decoder $f_{\text{dec}}$ and reconstructed image $\hat{\mathbf{x}}$ approximating $\mathbf{x}$).
Next, gradient-based attacks like FGSM \cite{fgsm} and PGD \cite{pgdattack} are performed against \cls in order to minimally manipulate $\mathbf{h}$ such that it resembles encoded representations of target image class $Z = z'$ (where $z' \neq z$, the original image class), yielding perturbed representation $\mathbf{h}'$.
Finally, $\mathbf{h}$ and $\mathbf{h}'$ are each fed into the VAE decoder to reconstruct corresponding output images $\hat{\mathbf{x}}$ and $\hat{\mathbf{x}}'$ (respectively),
where $\hat{\mathbf{x}}$ resembles input image class $Z = z$ and $\hat{\mathbf{x}}'$ resembles target class $Z = z'$.

\begin{figure*}
\centering
    \includegraphics[width=0.8\textwidth]{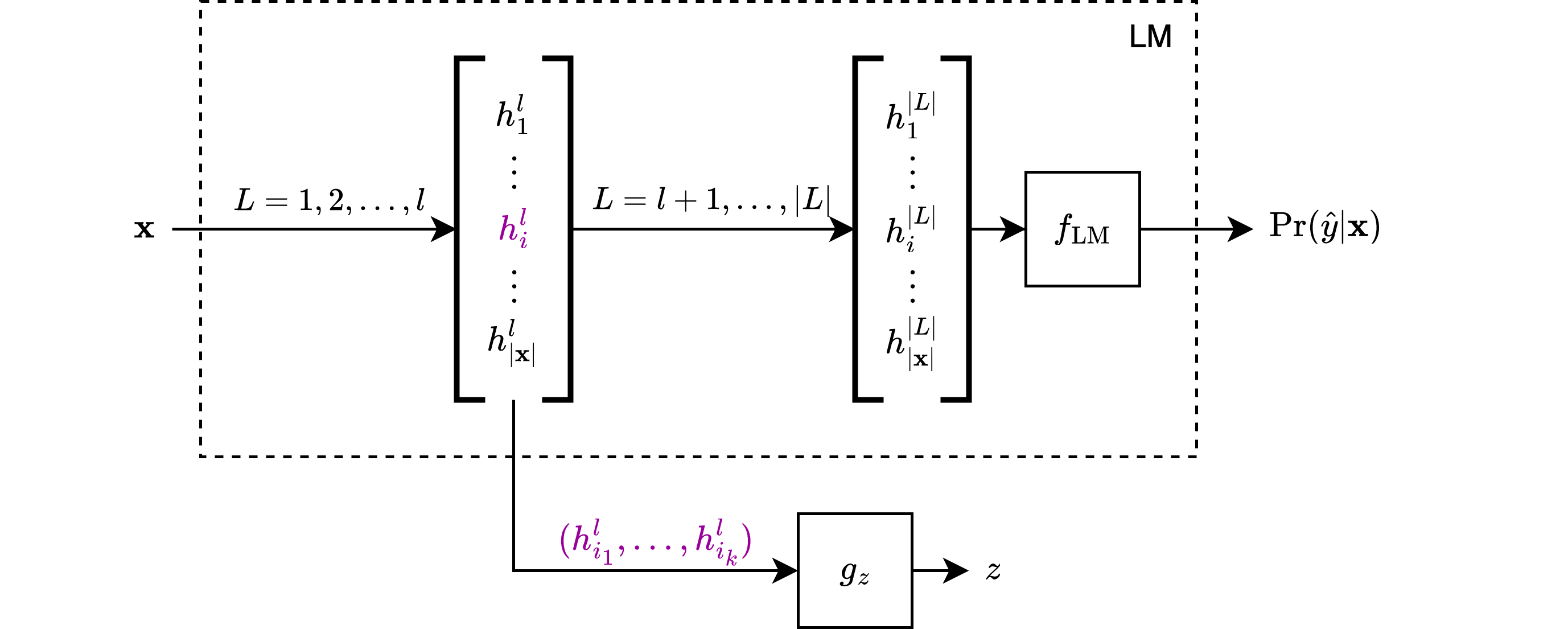}
    \caption{\textbf{Gradient-Based Interventions.} Input tokens $\mathbf{x} = (x_1, ..., x_{|\mathbf{x}|})$ are passed through layers $L = 1, ..., l$, where embedding $\mathbf{h}_i^l$ (encoding the value $Z = z$) is extracted from layer $l$ and given to \cls as input. 
    Next, the embedding is modified by gradient-based attacks on \cls to encode the counterfactual value $Z = z'$, then fed back into subsequent layers $L = l+1, ..., |L|$ 
    and language modeling head $f_{\text{LM}}$ to obtain the intervened predictions $M(\mathbf{x} | \doop(Z = z'))$.
    }
    \label{fig:gbi}
\end{figure*}

We reformulate this approach in the context of causal probing as visualized in \cref{fig:gbi}, treating layers $L = 1, ..., l$ as the encoder and layers $L = l+1, ..., |L|$ (composed with language modeling head $f_{\text{LM}}$) as the decoder, allowing us to target representations of property $Z$ across embeddings $\mathbf{h}_i^l$ of token $x_i \in \mathbf{x}$ in layer $l$.
We train \cls to predict $Z$ from a set of such $\mathbf{h}_i^l$,
then attack \cls using FGSM and PGD to intervene on $\mathbf{h}_i^l$ (representing the original value $Z = z$), producing $\mathbf{h}_i^{l^\prime}$ (representing the counterfactual value $Z = z'$).
Finally, we replace $\mathbf{h}_i^l$ with $\mathbf{h}_i^{l^\prime}$ in the LLMs' forward pass from layers $L = l+1, ..., |L|$, simulating the intervention $\doop(Z = z')$, and observe the impact on its word predictions $M(\mathbf{x} | \doop(Z = z'))$.
(See \cref{apx:gbi} for further details.)

\paragraph{Benefits and Drawbacks}
The key advantage of gradient-based interventions (GBIs) as a causal probing methodology is that they may be applied to any differentiable probe. For example, if we are investigating the hypothesis that $M$'s representation of $Z$ is captured by a linear subspace of representations in a given layer (cf. \citealp{linearsubspacehypothesis}), then we may train a linear probe and various nonlinear probes on representations and observe whether GBIs against the linear probe have a comparable impact to those against the nonlinear probes.
Alternatively, if we believe that a probe's architecture should mirror the architecture of the model it is probing (as argued by \citealp{pimentel2022bottleneck}), we may implement probes as such.
Finally, where previous intervention methodologies for causal probing have focused on \emph{nullifying} interventions that remove the representation of the target property $Z$ \cite{inlp,kernel_inlp,ravfogel2022linear,spectral_nonlinear,belrose2024leace}, GBIs allow one to perform targeted interventions that set LLMs' representations to counterfactual values $\doop(Z = z')$, effectively simulating the model's behavior under counterfactual inputs, which may be useful for predicting behaviors under various distribution shifts (see \cref{apx:iia}).
However, the benefits associated with GBIs do come with some important limitations, as we discuss in \cref{apx:gbi_limitations}.

\section{Experiments}\label{sec:experiments}
Our primary goal in the following experiments is to develop and test an experimental implementation of \calm using GBIs in the context of comparatively small, well-studied models and tasks in order to validate whether \calm can explain behavioral findings of earlier work in this simplified environment.
(We motivate this choice in greater detail in \cref{apx:whysimple}.)
Thus, we begin by examining BERT \cite{bert} and \rb \cite{roberta},\footnote{
    Specifically, \href{https://huggingface.co/bert-base-uncased}{\texttt{BERT-base-uncased}} and \href{https://huggingface.co/roberta-base}{\texttt{RoBERTa-base}} \cite{transformerslib}.
} two language models which have already been extensively studied in the context of probing
\cite{bertology,inlp,probeovertime,amnesic,causalgrammaticalnumber}.

\paragraph{Tasks}\label{sec:tasks}

We use the collection of 14 lexical inference tasks included in the ConceptNet \cite{conceptnet} subset of LAMA \cite{lama},
each of which are formulated as a collection of cloze prompts \citep{prompt}.
For example, the LAMA ``IsA'' task contains $\sim$2K hypernym prompts corresponding to the ``IsA'' ConceptNet relation (including, e.g., ``A laser is a \mask which creates coherent light'', where the task is to predict that the \mask token should be replaced with ``device'', a hypernym of ``laser''), with the remaining 13 LAMA ConceptNet tasks corresponding to other lexical relations such as ``PartOf'', ``HasProperty'', and ``CapableOf''. (See \cref{apx:tasks} for additional details.)

Using these task datasets allow us to 
test how the representation of each relation is used across all other tasks.
In the context of a single task $\gT_j$, intervening on a model's representation of the task-causal relation $Z_j$ allows us to measure the extent to which its predictions are attributable to its representation of the causal property $\rmZ_c = \{ Z_j \}$ (where a large impact indicates competence).
On the other hand, intervening on the representations of the other 13 lexical relations $Z_k \in \rmZ_e$ allows us (in the aggregate) to measure how much the model is performing task $\gT_j$ by leveraging representations of general, non-causal lexical information (where a large impact indicates incompetence).\footnote{
    Note that this experimental formulation makes the simplifying assumption that each environmental property is equally (un)related to the target property, which is not necessarily true; see \cref{apx:task_independence}.
}

\paragraph{Experimentally Measuring Competence}

Given LLM $M$ and task $\gT$, measuring the empirical competence $\hat\ct(M | \ggt)$ of $M$ given $\ggt$ requires us to specify an experimental model $E = (\rmZ, \ggt, S)$, where $\rmZ$ is a set of properties, $\ggt$ is a competence graph for task $\gT$, and $S$ is a scoring function that compares the predictions of $M$ and $\ggt$.
Given that each task $\gT_i$ is defined by a single causal lexical relation $Z_i$ (i.e., $\rmZ_{c_i} = \{ Z_i \}$), we model settings $\rvz$ as a collection of values $Z_j = z_j$ taken by each property $Z_j$ in the context of a specific task instance $(\rvx, \rvy) \sim \gT_i$,
where $Z_j = 1$ if $i = j$ (i.e., where the property $Z_j$ is the causal property for the task $\gT_i$) or $Z_j = 0$ otherwise. 
That is, for each instance $(\rvx, \rvy) \sim \gT_i$, the corresponding setting $\rvz$ is a one-hot vector whose $i$-th element $z_i = 1$.
We may specify $\gG_{\gT_i}$ in a similar manner: for task $\gT_i \sim P(\gX, \gY)$, outputs $\rvy \in \gY$ are causally dependent on the property $Z_i$
and invariant to other concepts $Z_j, j \neq i$.,
meaning that the only direct parent node of $\mathbf{y}$ in $\gG_{\gT_i}$ is $Z_i$.
Finally, as we are dealing with masked language models whose output space $\gY$ for each task consists only of single tokens in $M$'s vocabulary $V_M$, our experimental model can define the scoring function $S$ as the overlap $\ovl(\rvy_i, \rvy_j)$ for top-$k$ token predictions $\rvy_i = \{ y_1, ..., y_k \} \subset V_M$, where $\ovl(\cdot, \cdot)$ is the size of the intersection of each set of predictions divided by the total number of predictions $\ovl(\rvy_i, \rvy_j) = \frac{| \rvy_i \cap \rvy_j |}{k}$, and 
$\hat\ct_k$ denotes the empirical competence $\hat\ct$ as measured using only the top-$k$ token predictions $\rvy_i$.
(See \cref{apx:experimentalmetric} for additional details on how we compute competence in each experiment.)

\paragraph{Probes}\label{sec:cls}
We implement probes \cls as a 2-layer MLP over each language model's final hidden layer, and train the probe on the task of classifying whether there is a particular relation $Z$ between a final-layer \mask token in the context of a cloze prompt 
and the final-layer object token from the ``unmasked'' version of the same prompt. All reported figures are the average of 10 runs of our experiment, using different randomly-initialized \cls each time. (See \cref{apx:cls} for further details.)

\paragraph{Interventions}\label{sec:ptbn}
We implement GBIs against \cls using two gradient attack strategies, FGSM \cite{fgsm} and PGD \cite{pgdattack}.
We bound the magnitude of each intervention as follows: where $h$ is the input to \cls and $h^{\prime}$ is the intervened representation following a GBI, $||h - h^{\prime}||_\infty \leq \epsilon$.
For all experiments reported in our main paper, we use FGSM with $\epsilon = 0.1$. (See \cref{apx:gbi} for more details and PGD results.)

\section{Results}\label{sec:results}

\begin{figure}
    \centering
    \begin{subfigure}{0.5\textwidth}
        \centering
        \includegraphics[width=0.8\linewidth]{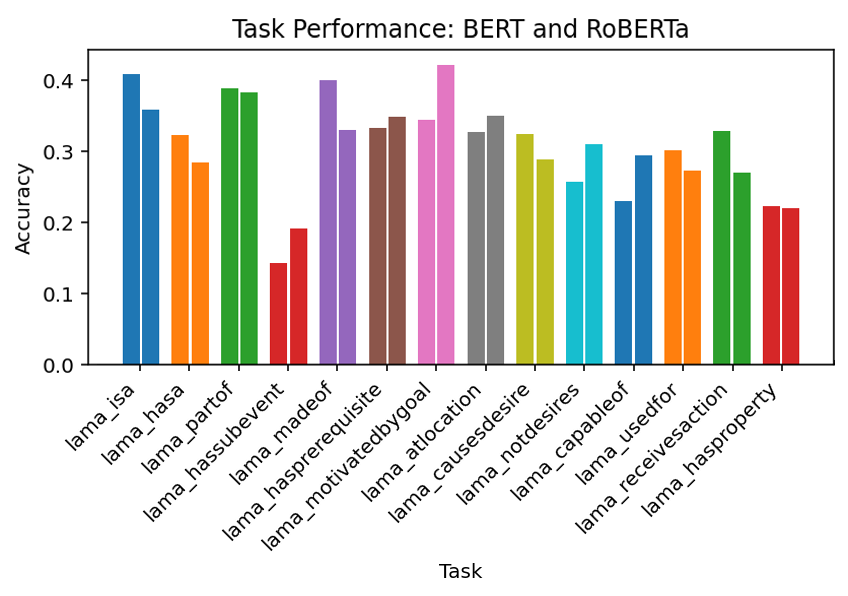}
    \end{subfigure}%
    \begin{subfigure}{0.5\textwidth}
        \centering
        \includegraphics[width=0.85\linewidth]{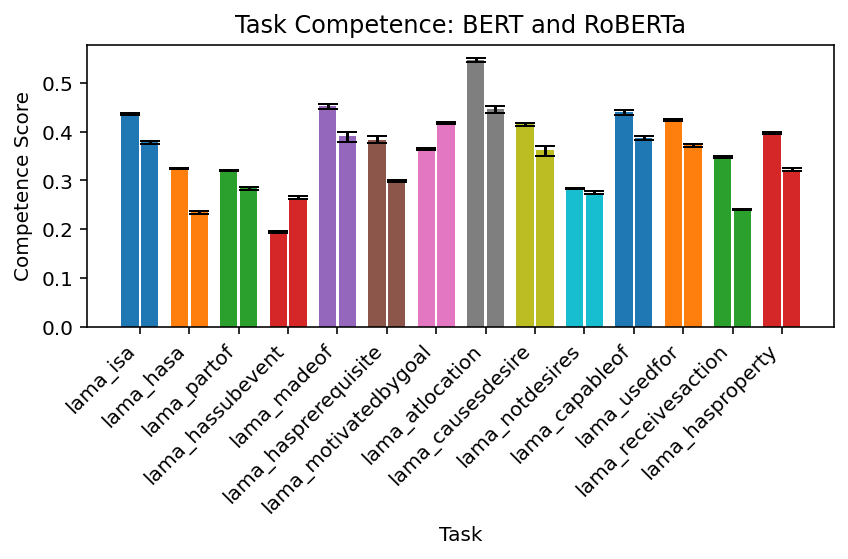}
    \end{subfigure}
    \caption{Performance (left) and competence (right) of BERT (left bars) and \rb (right bars) for all tasks, using FGSM with $\epsilon = 0.1$. In the competence plot, y-values are the average competence score and error bars are the maximum and minimum competence score, as measured over 10 experimental iterations (each with a different randomly-initialized probe $g_Z$).}
    \label{fig:mainresult}
\end{figure}

In \cref{fig:mainresult}, we visualize the performance and competence of BERT and \rb across the test set of each LAMA ConceptNet task.
Performance is measured using $(0, 1)$-accuracy, competence is measured using the experimental competence metric in \cref{eqn:experimental}, and both metrics are averaged across the top-$k$ predictions of each model for $k \in [1, 10]$. 
That is, for ground truth $(\mathbf{x}, y)$ and $n = 10$, we compute accuracy and competence as follows:
\[ 
    \acc(M) = \frac{1}{n} \sum_{k=1}^n \mathbbm{1} [y \in \topk_{\hat{y}} \Pr_M(\hat{y}|\mathbf{x})]
    \quad\quad \text{and} \quad\quad 
    \hat\ct(M | \ggt) = \frac{1}{n} \sum_{k=1}^n \hat\ct_k(M | \ggt)
\]
To account for stochasticity in initializing and training probes $g_Z$, scores are also averaged over 10 randomized experiments for each target task where the probe is randomly re-initialized each time (resulting in different GBIs).

\subsection{Analysis}\label{sec:analysis}
\paragraph{Performance}
While their accuracies on individual tasks vary, BERT and \rb have quite similar aggregate performance: BERT outperforms \rb on just over half (8/14) of the tasks, achieving essentially equivalent performance when averaged across all tasks (0.3099 versus 0.3094). 

\paragraph{Competence}
Given our experimental model $E$ with $m = 14$ tasks, consider a random baseline language model $R$ whose predictions always change in response to each intervention, making equal use of all properties in each task. $R$ would yield a competence score of $\gC(R | \ggt) = \frac{1}{m} \approx 0.0714$ for each task.
Both BERT and \rb score well above this threshold for all tasks, meaning that they are much more competent than a model that does not distinguish between causal and environmental properties.
\rb is consistently less competent than BERT (on 12/14 tasks), and also has lower competence scores averaged across all tasks (0.381 vs. 0.334);
but for the two tasks where \rb is more competent than BERT, it also achieves substantially higher performance. 
More generally, relative performance and competence are correlated: the Spearman's Rank correlation coefficient between the average difference in accuracy and average difference in performance is a moderately strong positive correlation $\rho = 0.508$ with significance $p = 0.064$.

\paragraph{Explananda}
The performance of BERT and \rb on lexical inference tasks such as hypernym prediction has been shown to be highly variable under small changes to prompts \cite{hypernym_prompt,systematicity,ettinger,pararel}.
Our findings offer a potential explanation for such brittle performance: BERT and \rb's partial competence in hypernym prediction indicates that it should be possible to prompt these models in a way that will yield high performance, but that its reliance on spurious lexical associations may lead it to fail when these correlations are broken -- e.g., by substituting singular terms for plurals \cite{systematicity} or paraphrasing a prompt \cite{pararel}.

\section{Related Work}\label{sec:relwk}

\paragraph{Causal Probing}
Most related to our work is amnesic probing \cite{amnesic}, as discussed in \cref{sec:causal}. 
\citet{causalgrammaticalnumber} applied amnesic probing to study the use of grammatical number representations in performing an English verb conjugation prompt task.
As this experiment involves intervening on the representation of a property which is causal with respect to the prompt task, it may be understood as an informal instantiation of \calm (albeit without considering environmental properties or measuring competence).

\paragraph{Gradient-based Interventions}\label{sec:relwk_gbi}
\citet{earlygbi} developed an approach similar to GBIs without explicit use of gradient-based adversarial attacks. Their methodology is equivalent to performing a targeted, unconstrained attack, where gradient updates are continually applied to embeddings until the target probe loss saturates (irrespective of perturbation magnitude).
In such attacks, it is standard practice to constrain the magnitude of resulting perturbations \cite{fgsm,pgdattack,latentattack}, which we do here in order to minimize the effect of ``collateral damage'' done by such attacks (see \cref{apx:gbi}).
Failing to impose such constraints may result in indiscriminate damage to representations \cite{canby2024cpr} (see \cref{apx:gbi_limitations} for further discussion).

\paragraph{Unsupervised Probing}
Instead of training supervised probes to predict a pre-specified property of interest (as we do here),
an alternative approach is to train \emph{unsupervised} probes such as Sparse Auto-Encoders (SAEs; \citealp{subramanian2018spine,yun2021dictionary,cunningham2023sae}), which learn a ``dictionary'' of features that can be used to sparsely represent embeddings,
and can also be used to control models' use of these learned features \cite{bricken2023monosemanticity,templeton2024scalingmonosemanticity}.
Unlike supervised probing, unsupervised dictionary features must be retroactively interpreted in order to determine their relationship to a given task \cite{davies2024cognitive}; but given a suitable approach to interpreting such features (see, e.g., \cite{bills2023language,paulo2024automatically}) and a sufficiently reliable method for intervening on them (cf. \cite{canby2024cpr}), it is also possible to implement \calm using unsupervised probes like SAEs.

\section{Conclusion}\label{sec:conclusion}
In this work, we introduced \calm, a general analysis framework that enables the study of LLMs' linguistic competence using causal probing, including the first quantitative measure of linguistic competence.
We developed the gradient-based intervention (GBI) methodology, a novel causal probing method that can target a far greater range of representations than previous techniques, expanding the scope of causal probing to new questions in LLM interpretability and analysis.
Finally, we carried out a case study of \calm using GBIs, analyzing BERT and \rb's competence across a collection of lexical inference tasks, 
finding that even a simple experimental model is sufficient to 
explain their behavior across a variety of lexical inference tasks.

\paragraph{Future Work}
While the simplified experimental setting considered in this work is an important first step in empirically validating our theoretical \calm framework, competence metric, and GBI methodology, we anticipate a much broader range of future research directions and potential applications for \calm.
For instance, the \calm framework could easily be deployed to study how various model training and fine-tuning choices impact learned representations (see \cref{apx:representation}),
or to characterize tasks based on mutual dependency structures
in order to potentially improve multitask learning
(see \cref{apx:multitask}) or
predict negative interactions between tasks used for model fine-tuning (see \cref{apx:dependencies}).
Finally, rather than comparing against a pre-specified ground truth task structure $\ggt$, it is also possible to discover a causal model describing an LLM's implicit task representation by combining \calm with traditional causal graph discovery algorithms (see \cref{apx:graphdiscovery}).

\newpage
\section*{Acknowledgments}

This work is supported in part by the National Science Foundation and the Institute of Education Sciences, U.S. Department of Education, through Award \#2229612 (National AI Institute for Inclusive Intelligent Technologies for Education). Any opinions, findings, and conclusions or recommendations expressed in this material are those of the author(s) and do not necessarily reflect the views of National Science Foundation or the U.S. Department of Education.

We thank Julia Hockenmaier, Marc E. Canby, Francesco Pinto, Bo Li, and Arindam Banerjee for helpful discussions regarding our framework and empirical analysis, as well as their valuable feedback on earlier drafts of this manuscript.

\printbibliography

\appendix

\section{Limitations}\label{sec:limitations}

\subsection{Gradient-Based Interventions}\label{apx:gbi_limitations}

For causal probing to operate successfully -- as is required to reliably deploy \calm in practice -- it is important that probes leverage the underlying model's representation of the target property to make predictions, rather than relying on spurious information.
However, there is some evidence that probes often leverage such spurious information \cite{probingsurvey,kumar2022probing,canby2024cpr}.
For instance, in followup work studying our GBI methodology alongside other causal probing methods, \citet{canby2024cpr} find that each method they studied (including GBIs) shows a tradeoff between its ability to manipulate the targeted property (\emph{completeness}) and the extent to which it also modifies the representation of other, non-targeted property (\emph{selectivity}).
Notably, they also found that the flexibility of GBIs allows for precise control this tradeoff by modulating the magnitude of perturbations ($\epsilon$), an advantage that is not shared by most other causal probing methods.

Furthermore, while GBIs are applicable to a more general range of model representations than most prior intervention methods (see \cref{sec:gbi}), this generality comes with a lack of constraints on probes (\cls); and as a result, GBIs cannot provide the strong theoretical constraints on collateral damage as can methods like, e.g., INLP \cite{inlp}, which provably preserves distances between embeddings as well as possible while completely removing the linear representation of the target property (which also generally leads to higher selectivity in practice \cite{canby2024cpr}).
To minimize collateral damage to representations, the magnitude of perturbations should be modulated via constraints on gradient attacks against \cls (see \cref{sec:ptbn}) and experimentally validated to control the damage done to representations (see \cref{apx:gbi}); and in the ideal case, should be calibrated to achieve the desired tradeoff between \emph{selectivity} and \emph{completeness} (a novel procedure introduced in \cite{canby2024cpr}, a followup work building on GBIs as initially developed in this work).
Alternatively, in cases where the structure of representations is believed to satisfy strong assumptions (e.g., being restricted to a linear subspace; \citealp{linearsubspacehypothesis}) or strong upper bounds on collateral damage are required, \calm interventions can be implemented with methods like INLP rather than GBIs.\footnote{
    It may also be possible to control for collateral damage by developing GBI strategies that offer more principled protection against damage to non-targeted properties, such as adding a loss term to penalize damage to non-targeted probes or leveraging interval bound propagation \cite{gowal2019intervalboundpropagation} to place intervened embeddings inside the adversarial polytope for non-targeted properties.
    We leave such possibilities to future work.
}

\subsection{Simple Experimental Setting}\label{apx:whysimple}
As noted in \cref{sec:experiments}, our primary goal in our experiments is to validate \calm by testing it in a simplified experimental setting consisting of comparatively small, well-studied models and tasks.
As such, we need models that are \emph{just complex enough} for \calm to be applicable (i.e., neural language models that are capable of performing the tasks we consider at a nontrivial level of performance), making BERT and \rb ideal candidates; 
and in future work plan to scale \calm to more complex contexts covering larger, more powerful models as they perform more difficult tasks (see \cref{apx:futurework}).
This is a common setting in the context of substantial recent interpretability work: first, a theoretical framework is developed for interpreting an internal representation or mechanism and initially tested in the context of ``toy'' models or tasks \cite{elhage2021circuit,olsson2022induction,zhong2023pizza,geiger2023causalinterp}, and subsequent work scales these frameworks to the context of larger models ``in the wild'' \cite{wang2023interpinthewild,conmy2023automated,wu2023boundlessdas}.
Analogously, all of our major contributions (the \calm framework, competence metric, and GBI causal probing method) are directly scalable to much larger, more recent LLMs (e.g., \citealt{zhang2022opt,bigscience2022bloom,llama,llama2,groeneveld2024olmo}, etc.).

\subsection{Task Independence}\label{apx:task_independence}

In our experiments, we modeled the 14 LAMA ConceptNet tasks as representing fully independent properties, which is not necessarily true -- e.g., knowing that a tree is made of bark or contains leaves tells us something about whether it is a type of plant. 
However, in the aggregate (with impacts summed across 14 widely-varying lexical relation types in computing the final competence score for each task; see \cref{apx:experimentalmetric}), it may nonetheless be appropriate to treat the relations which are not causal with respect to a given task as collectively capturing spurious lexical associations.

\section{Experimental Details}\label{apx:suppres}
\subsection{Tasks}\label{apx:tasks}
The full set of LAMA ConceptNet tasks is as follows: IsA, HasA, PartOf, HasSubEvent, MadeOf, HasPrerequisite, MotivatedByGoal, AtLocation, CausesDesire, NotDesires, CapableOf, UsedFor, ReceivesAction, and HasProperty.
We split each task dataset into train, validation, and test sets with a random $80\%/10\%/10\%$ split. Train and validation instances are fed to each model to produce embeddings used to train \cls and select hyperparameters, respectively; and test instances are used to measure LLMs' competence with respect to each task by observing how predictions change under various interventions.
In all experiments, we restrict each model $M$'s output space for each task $\gT$ to the subset of vocabulary $V_M$ that occurs as a ground-truth answer $y^*$ for at least one instance $(\rvx, y^*) \sim \gT$ in the respective task dataset. This lowers the probability of false negatives in evaluation (e.g., penalizing the model for predicting $\hat{y} =$ ``mammal'' for ``a dog is a type of $y$'' instead of $y^* =$ ``animal'').

\subsection{Probes}\label{apx:cls}
We use BERT's final layer $L$ to encode \hil embeddings for each such example, where $i$ is the index of the \mask token or target word in the input prompt $x_i$. 
To encode the \mask token, we issue BERT masked prompts (as discussed above) to extract $h_{\mask}$, then repeat with the \mask token filled-in with the target word to encode it as $h_+$ (e.g., ``device'' in ``A laser is a device which creates coherent light.''),
and concatenate matching embeddings $h = (h_{\mask}; h_+)$ to produce positive ($y = 1$) training instances. We also construct one negative ($y = 0$) instance, $h = (h_{\mask}; h_-)$, for each $h_{\mask}$ by sampling an incorrect target word $x_i$ corresponding to an answer to a random prompt from the same task, feeding it into the cloze prompt in the place of the correct answer,
and obtaining BERT's contextualized final-layer embedding of this token ($h_-$).
Finally, we train \cls on the set of all such $(h, y)$.

We implement \cls as a multi-layer perceptron with 2 hidden layers, each with a width of 768 (which is one half the concatenated input dimension of 1536), using ReLU activations and dropout with $p=0.1$, training it for 32 epochs using Binary Cross Entropy with Logits Loss\footnote{\url{https://pytorch.org/docs/stable/generated/torch.nn.BCEWithLogitsLoss.html}} 
and the Adam optimizer, saving the model from the epoch with the highest validation-set accuracy for use in all experiments.

For all competence results reported in \cref{sec:results}, we run the same experiment 10 times -- each with a different random initialization of \cls and shuffled training data -- and report each figure as the average among all 10 runs.

\begin{figure}
    \centering
    
    \begin{subfigure}{0.5\textwidth}
        \centering
        \includegraphics[width=0.95\linewidth]{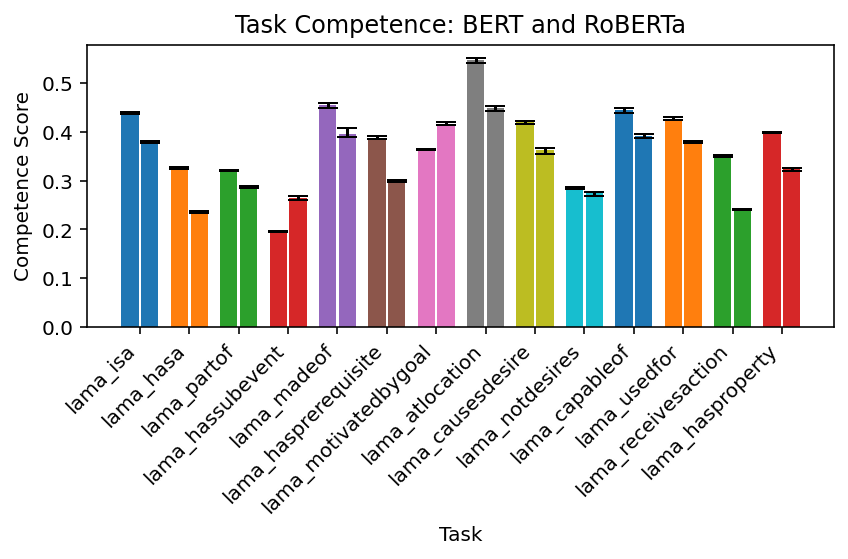}
    \end{subfigure}
    \caption{Competence of BERT (left bars) and \rb (right bars) for all tasks, using PGD with $\epsilon = 0.1$. Y-values are the average competence score and error bars are the maximum and minimum competence score, as measured over 10 experimental iterations (each with a different randomly-initialized probe $g_Z$).}
    \label{fig:pgdresult}
\end{figure}

\subsection{Interventions}\label{apx:gbi}
For embedding $h$, target (counterfactual) class $y'$, probe \cls, loss function $\mathcal{L}$, and $L_{\infty}$-bound $\epsilon \in \{ 0.01, 0.03, 0.1, 0.3 \}$\footnote{
    All reported results use $\epsilon = 0.1$, as greater $\epsilon$ resulted in unacceptably high ``collateral damage'' across target tasks (e.g., even random perturbations of magnitude $\epsilon = 0.3$ do considerable damage), and lesser values meant that predictions changed on target tasks consisted of only a few test instances.
}, 
each intervention (gradient attack) $g_{z}$ may be used to produce perturbed representations $h^\prime= g_{z}(h, y', f_{\text{cls}}, \mathcal{L}, \epsilon)$ where $||h - h^{\prime}||_\infty \leq \epsilon$.
In particular, given $h = (h_\mask; h_\pm) \in \mathbb{R}^{2d}$, let $h_\mask^\prime$ be the first $d$ dimensions of $h^{\prime}$ (which also satisfies the $L_{\infty}$-bound with respect to $h_\mask$, $||h_\mask - h_\mask^\prime||_\infty \leq \epsilon$).
To measure BERT's use of internal representations of $Z$ on each prompt task, we evaluate its performance when perturbed $h_\mask^\prime$ is used to compute masked-word predictions, compared to unperturbed $h_\mask$.

Our intent in intervening only on the final-layer mask embedding $h_\mask$ in our experiments is that, in the final layer of a masked language model such as BERT or \rb, the only embedding which is used to compute masked-word probabilities is that of the \mask token. Thus, any representation of the property that is \emph{used} by the model in its final layer must be a part of its representation of the \mask token, preventing ``recoverability'' phenomena such as those observed by \citet{amnesic}.

\paragraph{FGSM}
FGSM \cite{fgsm} takes one gradient step of magnitude $\epsilon$ in the direction that minimizes the loss of a classifier (here, the probe $f_{\text{cls}}$) with respect to target class $y'$.
We implement FGSM interventions as
\[
    h^\prime = h + \epsilon \cdot \text{sgn} ( \nabla_{h} \mathcal{L}(f_{\text{cls}}, x, y') )
\]
where $\gL$ is the same loss function used to train $f_{\text{cls}}$ (here, binary cross entropy).

\paragraph{PGD}
PGD \cite{pgd,pgdattack} iteratively minimizes the loss of a classifier (here, the probe $f_{\text{cls}}$) with respect to target class $y'$ by performing gradient descent within a $L_\infty$ ball of radius $\epsilon$.
We implement PGD interventions as $h^\prime = h^T$, where
\[
    h^{t+1} = \Pi_{N(h)} \big( h^t + \alpha \cdot \text{sgn} ( \nabla_{h} \mathcal{L}(f_{\text{cls}}, x, y) ) \big)
\]
for iterations $t = 0, 1, ..., T$, projection operator $\Pi$, $L_{\infty}$-neighborhood $N(h) = \{ h^{\prime} : ||h - h^{\prime}|| \leq \epsilon \}$, and $\gL$ is the same loss function used to train $f_{\text{cls}}$ (here, binary cross entropy).
This method also introduces two hyperparameters: the number of PGD iterations $T$ and step size $\alpha$. 
We use hyperparameter grid search over $\alpha \in \{ 0.001, 0.003, 0.01, 0.03 \}$
and $T \in \{ 20, 40, 60, 80, 100 \}$, finding that setting $\alpha = \frac{\epsilon}{10}$ and $T = 40$ produces the most consistent impact on \cls accuracy across all tasks; so we use these values for the results visualized in \cref{fig:pgdresult}.

\subsection{Compute Budget}
BERT-base-uncased has 110 million parameters, and RoBERTa-base has 125M parameters.
As our goal is to study the internal representation and use of linguistic properties in existing pre-trained models, and we are not directly concerned with training or fine-tuning such models, we use these models only for inference (including encoding text inputs, using embeddings to train probes, and feeding intervened embeddings back into the language models). 
The only models we trained were probes $g_Z$, which each had 1.77M parameters.

Each experimental iteration (including encoding text inputs, training probes on all 14 tasks, and performing all GBIs) for either BERT or RoBERTa took less than one hour on a single NVIDIA GeForce GTX 1080 GPU, meaning that running all 10 iterations across both language models took less than 20 hours on a single GPU. Each iteration, probe, and GBI can easily be parallelized across GPUs: in our case, running all iterations across both models took less than 3 hours total across 8 GTX 1080 GPUs.

\section{Competence Metric}\label{apx:metric}

\subsection{Comparison With IIA}\label{apx:iia}
As noted in \cref{sec:metric}, the $\ct(M | \ggt)$ metric defined in \cref{eqn:competence} is an adaptation of the Interchange Intervention Accuracy (IIA) metric \citep{iia,geiger2023causalinterp},
which evaluates the faithfulness of a causal abstraction like $\ggt$ as a (potential) explanation of the behavior of a ``black box'' system like $M$.
In our case, this is equivalent to evaluating the competence of $M$ on task $\gT$, provided that $\ggt$ is the appropriate SCM for $\gT$, as an LLM is competent only to the extent that its behavior is determined by a causally invariant representation of the task.\footnote{
    For many tasks, there is more than one valid $\ggt$ (see, e.g., the ``price tagging game'' constructed by \citet{wu2023boundlessdas}). In such cases, $\ct(M | \ggt)$ should be computed with respect to each valid $\ggt$ and the highest result should be selected, as conforming to any such $\ggt$ carries the same implications.
}
IIA requires performing \emph{interchange interventions} $\doop_{II}(\rvz_j)$, where the part of $M$'s intermediate representation of input $\rvx_i$ hypothesized to encode latent variables $\rmZ$ (taking the values $\rvz_i$ when provided input $\rvx_i$) is replaced with 
that of $\rvx_j$ (which, in principle, means that the modified representation encodes the values $\rvz_j$ instead of $\rvz_i$),
at which point these interchange interventions are used to compute predictions $M(\rvx_i | \doop_{II}(\rvz_j))$, and the output is compared with $\ggt(\rvx_i | \doop(\rvz_j))$ to measure how faithfully $\ggt$ predicts $M$'s behavior under these interventions.
Thus, given access to high-quality interchange interventions over $M$, IIA measures the extent to which $\ggt$ correctly models $M$'s behavior under counterfactuals, and thus its faithfulness as a causal abstraction of $M$.

To adapt IIA to the context of causal probing and define $\ct(M | \ggt)$, we replace instance-level interchange interventions $\doop_{II}$ with concept-level interventions $\doop_Z$ for any given property $Z$. That is, instead of swapping $M$'s representation of variables $\rmZ = Z_1, ..., Z_k$ given input $\rvx_i$ with that of $\rvx_j$, we intervene on the representation of each property $Z \in \rmZ$ at the level of arbitrary values $\rvz: Z_1 = z_1, ..., Z_k = z_k$ that need not correspond to previously observed $\rvx$, allowing us to simulate the behavior of $M$ under previously-unseen distribution shifts (i.e., settings $\rvz$ representing previously-unseen combinations of values) and in doing so predict $M$'s consistency with a given causal model $\ggt$ under these new conditions.
As one of our primary motivations in studying LLM competence is to provide a framework useful for predicting and explaining behavior under distribution shifts, $\ct$ is more appropriate than IIA in this setting.
However, this also introduces greater room for error: where interchange interventions only requires modifying representations to match the values taken by another input -- as counterfactual representations can be obtained simply by ``plugging in'' representations from a different input -- computing $\ct$ instead requires one to perform open-ended interventions that may not correspond to any ground-truth input, in which case there may be no region of the embedding space that corresponds to the intended setting $\rvz$ \cite{geiger2020neural,abraham2022cebab,canby2024cpr}.

\subsection{Experimental Competence Metric}\label{apx:experimentalmetric}

To compute the expectation in \cref{eqn:competence} for test set $\{ \rvx_i, \rvy_i, \rvz_i \}_{i=1}^n \sim \gT \times \rmZ$, we sum the competence score over all samples $\rvx_i$ and perform one intervention $\doop(Z_j = 0)$ corresponding to each concept $Z_j \in \rmZ$.\footnote{
    Note that this intervention changes the prediction $\ggt(\rvx_i) \neq \ggt(\rvx_i | \doop(Z_j = 0))$ if and only if $(\rvx_i, \rvy_i) \in \gT_j$ -- i.e., where the corresponding $(\rvz_i)_j = 1$ -- otherwise, $(\rvz_i)_j$ is already $0$, so the intervention has no effect. Thus, as $\ct(M | \ggt)$ measures $M$'s consistency with $\ggt$, then to the extent that $M$ is competent, its prediction should change under all and only the same interventions as $\ggt$.
} 
As our goal is to measure the extent to which $M$'s behavior is attributable to an underlying representation of the causal property $Z_c$ or environmental property $Z \in \rmZ_e$, our experimental model defines $\ggt$'s predictions with reference to $M$'s original predictions $M(\rvx_i) = \hat{\rvy_i}$, according to the following principle: if $M$ is competent, then its prediction $M(\rvx_i) = \hat{\rvy_i}$ is wholly attributable to its representation of causal property $Z_c$, so its predictions $M(\rvx_i | \doop(Z_c)) = \hat{\rvy_i}'$ will not overlap with its original predictions $\hat{\rvy_i}$ (i.e., $\ovl(\hat{\rvy_i}, \hat{\rvy_i}') = 0$); and conversely, a competent $M$ will make the \emph{same} predictions $M(\rvx_i | \doop(Z_j)) = \hat{\rvy_i}''$ for any $Z_j \in \rmZ_e$, because its prediction is not caused by its representation of these environmental properties (i.e., $\ovl(\hat{\rvy_i}, \hat{\rvy_i}'') = 1$).
Motivated by this reasoning, our experimental model defines $\ggt(\rvx_i | \doop(Z_j = 0)) = M(\rvx_i)$ for environmental $Z_j \in \rmZ_e$; and for causal property $Z_c$, defines $\ggt(\rvx_i | \doop(Z_c = 0)) = \{ y' \in V_M : y' \notin M(\rvx_i) \}$ (i.e., the set of all tokens $y'$ in $M$'s vocabulary that were not in its original prediction $M(\rvx_i)$).
Thus, under experimental model $E$, we approximate $\ct(M | \ggt)$ by computing it as follows:

\begin{align}\label{eqn:experimental}
    \hat\ct(M | \ggt) = \frac{1}{n \cdot m} \sum_{i=1}^n \sum_{j=1}^m \ovl \Big( M \big( \rvx_i | \doop(Z_j = 0) \big), \ggt \big( \rvx_i | \doop(Z_j = 0) \big) \Big)
\end{align}

Notably, our experimental model $E$ only accounts for the relationship between $M$'s intervened and non-intervened predictions, independently of ground truth labels -- instead, what is being measured is $M$'s consistency under meaning-preserving interventions $\doop(Z_{j^\prime})$ and its mutability under meaning-altering interventions $\doop(Z_{j})$.
However, as we find in \cref{sec:analysis}, the resulting competence metric $\ct(M | \ggt)$ is nonetheless useful for predicting $M$'s accuracy.

\section{Future Work}\label{apx:futurework}
\subsection{Representation Learning}\label{apx:representation}
The \calm framework, competence measure, and GBI methodology developed in \cref{sec:calm,sec:gbi} are sufficiently general to be directly applied to analyze arbitrary LLMs on any language modeling task whose causal structure is already well understood (or, for tasks where this is not the case, we may apply the causal graph discovery approach described in \cref{apx:graphdiscovery}), allowing us to study the impact of various model architectures, pre-training regimes, and fine-tuning strategies on the representations LLMs learn and use for arbitrary tasks of interest.

\subsection{Multitask Learning}\label{apx:multitask}
Are high competence scores on task $\gT$ correlated with an LLMs' robustness to meaning-preserving transformations (see, e.g., \citealp{pararel}) on tasks $\gT'$ that share several causal properties $\rmZ_c$ with task $\gT$.
Through the lens of causally invariant prediction \citep{icp,irm,invariance}, this hypothesis is likely true (however, see \citealt{risksofirm} for appropriate caveats) -- if so, this would make it possible to use clusters of related tasks to predict LLMs' robustness (and other behavioral patterns, such as brittleness in the face of distribution shifts introduced by spurious dependencies) between related tasks using \calm, given an appropriate experimental model.
Furthermore, the ability to characterize tasks based on mutual (learned) dependency structures could be valuable in transfer learning applications such as guiding the selection of auxiliary tasks in multi-task learning \citep{multitask} or predicting the impact of intermediate task fine-tuning on downstream target tasks \citep{intertraining}.

\subsection{Task Dependencies}\label{apx:dependencies}
Another possible application of \calm concerns causal invariance under multi-task applications.
Existing approaches in invariant representation learning generally require task-specific training \cite{invarreplrn}, as the notion of invariance is inherently task-centric (i.e., the properties which are invariant predictors of output values vary by task, and different tasks may have opposite notions of which properties are causal versus environmental),
so applying such approaches 
to train models to be causally invariant with respect to a specific downstream task $\gT$ is expected to come at the cost of performance on other downstream tasks $\gT'$.
Therefore, considering the recent rise of open-ended, task-general LLMs \cite{zhang2022opt,bigscience2022bloom,llama,llama2,groeneveld2024olmo},
it is important to understand the relationship between different task dependencies learned when fine-tuning task-general models on specific downstream tasks to account for applications involving tasks with different (and perhaps contradictory) causal structures, such as \calm.

\subsection{Causal Competence Graph Discovery}\label{apx:graphdiscovery}
A key feature of \calm is that, instead of simply measuring consistency with respect to a known, static task description $\ggt$, the competence metric in \cref{eqn:competence} can also be used to dynamically discover a competence graph $\gG$ which most faithfully explains a model $M$'s behavior in a given task or context (see \cref{sec:metric}) by computing $\gC(M | \gG)$ ``in-the-loop'' of existing causal graph discovery algorithms like IGSP \citep{cgdiscovery}. 
Such algorithms could be used to suggest likely competence graphs based on interventional data collected by running \calm experiments, to recommend the experiments that would yield the most useful interventional data for the graph discovery algorithm, or to evaluate candidate graphs $\gG$ according to $\gC(M | \gG)$, terminating the graph discovery algorithm once a competence graph $\gG$ that offers sufficiently faithful explanations of $M$'s behavior has been found (e.g., where $\gC(M | \gG) > \tau$ for some threshold $\tau$).
In this case, it is still necessary to define the set of properties $\rmZ$ being probed and the scoring function $S$ used to compare the predictions of $M$ and $\gG$; but no knowledge of the causal dependencies (or structural functions $F: \pa(Z_j) \mapsto Z_j$ mapping from causal parents $\pa(Z_j)$ to causal dependents $Z_j$; see \citealt{bongers2021scm}) is required.

\end{document}